\documentclass[letterpaper]{article}
\usepackage[preprint]{aaai2027}
\usepackage[hyphens]{url}
\usepackage{graphicx}
\usepackage{natbib}
\usepackage{caption}
\usepackage{algorithm}
\usepackage{algorithmic}
\usepackage{makecell}
\usepackage{newfloat}
\usepackage{listings}
\DeclareCaptionStyle{ruled}{labelfont=normalfont,labelsep=colon,strut=off}
\floatstyle{ruled}
\newfloat{listing}{tb}{lst}{}
\floatname{listing}{Listing}

\usepackage{booktabs}

\usepackage[utf8]{inputenc}
\usepackage[T1]{fontenc}      
\usepackage{amsfonts}     
\usepackage{nicefrac}    
\usepackage{microtype}    
\usepackage{xcolor}     

\usepackage{amsmath}
\usepackage{amssymb}
\usepackage{amsthm}
\usepackage{mathtools}
\usepackage{array}
\usepackage{tabularx}

\usepackage{textcomp}
\usepackage{verbatim}
\usepackage{relsize}

\usepackage{arydshln}
\usepackage{multirow}
\usepackage{enumitem}
\usepackage{tikz}
\usepackage{threeparttable}

\newcommand{\arxivauthorsize}{\fontsize{11.7pt}{13pt}\selectfont}

\title{Training-Free Token-Level Steering for LLM Personalized Co-Writing}

\author{
    {\arxivauthorsize
    Wenhao Mao\textsuperscript{\rm 1},
    Chengbin Hou\textsuperscript{\rm 2}\corresponding,
    Weixiao Wang\textsuperscript{\rm 1},
    Jialiang Zhu\textsuperscript{\rm 3},
    Min Liu\textsuperscript{\rm 3},
    Yibin Hao\textsuperscript{\rm 3}\corresponding,
    Hairong Lv\textsuperscript{\rm 1}\corresponding
    }
}
\affiliations {
    \textsuperscript{\rm 1}Tsinghua University\\
    \textsuperscript{\rm 2} Fuyao University of Science and Technology\\
    \textsuperscript{\rm 3}Henan Provincial People's Hospital\\
    \url{wmao1701@gmail.com},~\url{houcb@fyust.edu.cn},~\url{lvhairong@tsinghua.edu.cn}
}
\begin{document}

\maketitle

\begin{abstract}
While Large Language Models (LLMs) show great promise for personalization, they often lack specialized domain knowledge.
Conventional solutions like fine-tuning struggle with high computational costs and rapid data updates, while Retrieval-Augmented Generation fails to provide fine-grained, token-level steering.
Furthermore, chat-based interfaces remain dominant, whereas productive co-writing paradigms have not yet been well exploited beyond the coding domain.
To this end, we introduce SteerWrite, a training-free framework designed for personalized co-writing.
Our method effectively adapts the base model to specialized domains without gradient updates, with specific designs tailored to small datasets.
Experiments demonstrate that SteerWrite achieves state-of-the-art performance across diverse datasets, metrics, and models, significantly reducing human editing effort.
\end{abstract}

\section{Introduction}
Large Language Models (LLMs) have garnered significant attention across academia and industry. Consequently, various institutions have released open-source models, such as the Qwen series \cite{yang2025qwen3} and Llama series \cite{meta2025llama4}.
Building upon these foundations, adapting models for specialized and personalized scenarios to construct domain-specific custom models has become a prevalent practice \cite{thirunavukarasu2023large,shool2025systematic,joel2024survey}.
To equip models with domain knowledge, fine-tuning is a common approach \cite{wu2025llm,joel2024survey,jeong2024fine}.
For instance, DragFT \cite{zheng2024fine} employs dictionary-enhanced methods and data-quality improvement strategies to boost fine-tuning effectiveness; STAF-LLM \cite{xu2025staf} uses multiple experts and task routers to adapt to the diverse task requirements.
However, real-world scenarios often face constraints such as limited computational resources, scarce domain datasets, and frequent updating needs due to evolving content \cite{thirunavukarasu2023large,joel2024survey,wu2025llm}.

Alternatively, Retrieval-Augmented Generation (RAG) \cite{lewis2020retrieval} offers a popular training-free solution by injecting relevant context into the prompt, requiring less computational overhead and allowing for rapid database updates \cite{komeili2022internet}.
For example, MedGraphRAG \cite{wu2024medical} enhances performance on downstream medical tasks by constructing knowledge graphs with a specialized retrieval mechanism; CBR-RAG \cite{wiratunga2024cbr} combines multiple embedding and retrieval strategies to bolster legal question-answering capabilities.
Despite the potential complexity of retrieval designs, RAG-based approaches typically possess only a single opportunity to inject retrieved content into the prompt (at some predefined locations) for a given response. As a result, this restricts RAG to prompt-level guidance, failing to offer the finer-grained (e.g., token-level) steering required during the generative process.

Furthermore, the interaction paradigm for domain-specific LLMs remains primarily turn-based dialogue (i.e., Chat).
Existing works mainly focus on constructing complex workflows to enhance performance through multi-turn Q\&A and human-AI interaction \cite{wang2023interactive, park2023generative}.
For instance, TAMA \cite{xu2025tama} adopts human-in-the-loop multi-turn interactions to assist doctors in comprehending clinical documents.
Nevertheless, such intricate processes significantly consume user time and patience, imposing a high cognitive load \cite{mysore2025prototypical}.
Moreover, some existing techniques prioritize achieving state-of-the-art results on static domain benchmarks rather than genuinely assisting domain practitioners in their actual workflows \cite{wiratunga2024cbr}.
In contrast, the software engineering domain has successfully adopted stream-based co-writing, like GitHub Copilot \cite{github_copilot2022GitHub} and OpenAI Codex \cite{openai_codex2025OpenAI}, where the model acts as a low-latency "shadow typist", significantly boosting developer productivity.
Surprisingly, this highly productive paradigm has not been well explored and exploited beyond the coding domain.

To address these challenges, we introduce SteerWrite, a locally deployable, training-free token-level framework designed for interactive co-writing in specialized domains.
Specifically, starting from the theoretical insight of model posterior probabilities, we leverage the pre-trained model's own internal representations to perform token-level retrieval and guidance tailored for the small dataset.
Multiple distribution calibration strategies are employed to ensure stable performance during generation.
This allows a local base model to be instantly converted into a personalized co-writing assistant using a small, external dataset without any gradient updates.

We evaluate SteerWrite on four diverse domain datasets and three Qwen3 model sizes, comparing it with seven training-free baselines. Additional experiments on Llama-3.2 models further verify its generalization across model families. Experimental results demonstrate that SteerWrite consistently outperforms competing methods across datasets, metrics, and architectures. Critically, it substantially reduces the edit distance \cite{lcvenshtcin1966binary} and user keystrokes required to achieve the desired content, while introducing only modest latency overhead for interactive co-writing. Apart from experiments and analysis, the main technical contributions are summarized as follows:
\vspace{-3pt}
\begin{itemize}[leftmargin=*]
\setlength{\itemsep}{1.5pt}
\setlength{\parskip}{1.5pt}
\item We provide a theoretical insight into the domain adaptation based on probability theory and Kernel Density Estimation (KDE) \cite{davis2011remarks}, demonstrating that an ideal personalized model can be effectively approximated solely through inference-time posterior probability correction.
\item We propose a lightweight and training-free framework that enables base models to achieve token-level personalized co-writing using small supplementary datasets.
\item We construct a comprehensive benchmark for interactive co-writing, including four domain datasets and an evaluation protocol for quantifying reductions in human editing effort. Our code and publicly releasable resources are freely available at \url{https://github.com/LengendaryHippopotamus/SteerWrite}.
\end{itemize}

\section{Related Work}
\subsection{Training-Free Adaptation}
\label{sec:Training-Free}
Pre-trained LLMs have demonstrated remarkable generalization capabilities in zero-shot settings \cite{brown2020language, wei2021finetuned}, enabling them to perform various tasks without explicit gradient updates.
To further enhance performance on specific tasks, In-context Learning \cite{dong2024survey} has been introduced, leveraging a few demonstrations within the prompt to guide the model's behavior.
Extending this paradigm to knowledge-intensive or domain-specific scenarios, RAG \cite{lewis2020retrieval, peng2025graph} has emerged as a dominant training-free solution. It retrieves relevant document chunks from an external corpus and appends them into the input context, effectively reducing hallucinations and incorporating up-to-date information \cite{gao2023retrieval}.

However, most RAG-based approaches operate at the prompt level, injecting context into the prefix to select a more favorable initialization point within the model's high-dimensional latent space before generation \cite{reynolds2021prompt,li2021prefix,qin2021learning}. 
While effective in some cases, this mechanism lacks granularity for open-ended generation tasks like co-writing, where the model requires continuous, step-by-step guidance to maintain specific stylistic or vocabulary constraints throughout the entire sequence.
As the generation sequence lengthens, the influence of the initial prompt may diminish or fail to correct local deviations, which is often referred to as "prompt decay" or "context drift" \cite{shi2024trusting,tian2024selective,dongre2025drift}.

To achieve finer-grained control, non-parametric approaches, such as $k$NN-LM \cite{khandelwal2019generalization}, attempt linear interpolation with retrieval distributions at the token level.
However, recent analysis reveals a critical disconnect between perplexity reduction and generation quality; such methods tend to yield significant probability gains only on specific sparse tokens while degrading the general distribution, rendering them unsuitable for open-ended generation \cite{wang2023knn}.
In contrast, SteerWrite transcends heuristic interpolation by grounding the steering process in a theoretical approximation of posterior probability.
By adapting this framework to modern LLMs with specialized distribution stabilization mechanisms, we achieve robust and fluent long-sequence generation that previous methods have failed to sustain.

\subsection{Stream-Based Co-Writing}
The paradigm of AI-assisted co-writing has witnessed revolutionary success in the domain of software engineering.
Tools like GitHub Copilot \cite{github_copilot2022GitHub} and OpenAI Codex \cite{openai_codex2025OpenAI} function as low-latency "shadow typists", utilizing stream-based completion to predict the subsequent code lines in real-time.
This interaction mode significantly reduces the cognitive load on developers by transforming the workflow from manual typing to review-and-edit \cite{ziegler2022productivity, bird2022taking}.

Despite this potential, such a productive stream-based paradigm has not been widely adopted in non-coding domains such as medical record writing, legal drafting, or personalized assistance.
Current interactions with domain-specific LLMs are mostly chat-based (e.g., ChatGPT), treating the model as a conversational partner rather than a writing assistant.
Research in human-AI interaction suggests that such turn-based interactions often differ significantly from the user's actual writing workflow, introducing friction and inefficiency \cite{mysore2025prototypical}.
Moreover, generating entire paragraphs in a single turn often requires substantial human post-editing to align with personal intent, increasing keystrokes and edit distances compared to the more granular and interactive completion suggestions \cite{lee2022coauthor,bhat2023interacting}.

This work aims to bridge this gap by enabling generalized base models to perform personalized and stream-based co-writing efficiently.
By minimizing the edit distance and aligning closely with user intent through token-level steering, we attempt to bring the productivity benefits of the "Copilot" experience to the broader textual domains.

\begin{figure*}[htb!]
  \begin{center}
    \vspace{-0.8em}
    \centerline{\includegraphics[width=0.99\linewidth]{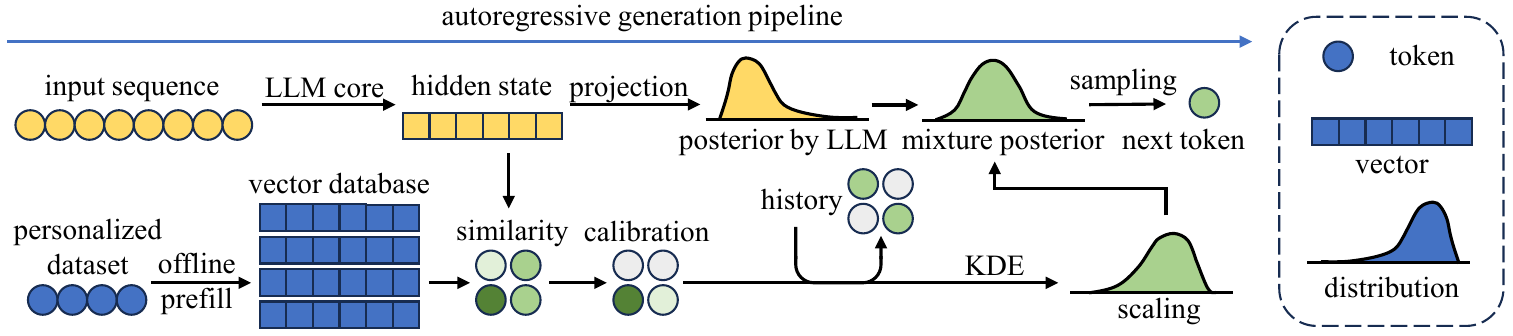}}
    \caption{Overall framework of SteerWrite.}
    \vspace{-1.8em}
    \label{fig:method_overview}
  \end{center}
\end{figure*}

\section{Theoretical Insights}
\label{sec:theory}
To explicitly guide the model toward optimal performance when incorporating newly introduced or personalized data, we formulate the following problem.

\textbf{Definition 3.1 Posterior Probability on Mixed Data.}
\label{def:nucleus}
Given a vocabulary list $\Omega$, a pre-trained model $f_1$ associated with a latent pre-training dataset $D_1$ (typically inaccessible), and a newly introduced dataset $D_2$, the posterior probability on the mixed dataset $D = D_1 \cup D_2$ can be defined as the probability of the next token given a token sequence $x_1 x_2 \cdots x_{t-1}$ with an assumed ideal model re-trained on $D$, which can be written as: 
\begin{equation}
\label{pxi}
   p(x_t \mid x_1 x_2 \cdots x_{t-1}, D)
\end{equation}

For notational convenience, let $x_{<t} = x_1 x_2 \cdots x_{t-1}$ denote the current input context. Furthermore, we may omit the subscript and refer to the given context as $x$ and the target token as $y$, i.e., $p(y\mid x)\triangleq p(x_t\mid x_{<t})$.
By applying Bayes' theorem, 
the posterior probability can be expanded as:
\begin{align}
    \label{pxmain}
    p(y \mid x) =& p(D_1 \mid x)p(y \mid x, D_1) + p(D_2 \mid x)p(y \mid x, D_2) \nonumber \\
    =& \frac{p(D_1)}{p(x)} \Big( p(x \mid D_1)p(y \mid x, D_1) \\
    &+ \frac{p(D_2)}{p(D_1)}p(x \mid D_2)p(y \mid x, D_2) \Big)
\end{align}

As the model $f_1$ has been pre-trained on $D_1$, we assume $f_1(y \mid x) \approx p(y \mid x, D_1)$.
In other words, $f_1$ has captured the posterior distribution of the dataset $D_1$.  
Consequently, the term $p(y \mid x, D_1)$ in Eq. (\ref{pxmain}) is given by $f_1$.
The priors ratio $p(D_2)/p(D_1)$ can be regarded as the ratio of the effective sizes of the datasets.

Note that $p(y \mid x, D)$ is a probability distribution over the token vocabulary $\Omega$, 
which satisfies the normalization condition:
\begin{equation}
    \sum_{r_k \in \Omega} p(y=r_k \mid x, D) = 1
    \label{sum}
\end{equation}
Therefore, we can eliminate the term $p(D_1)/p(x)$ in Eq. (\ref{pxmain}) via normalization. Furthermore, due to the autoregressive generative process, the likelihood of the context can be decomposed as:
\begin{equation}
\setlength{\abovedisplayskip}{2pt}
\setlength{\belowdisplayskip}{2pt}
    p(x_{<t} \mid D_k) = \prod_{j=1}^{t-1} p(x_j \mid x_{<j}, D_k)
\end{equation}

Therefore, provided that we can estimate $p(y \mid x, D_2)$, we can derive $p(y \mid x, D)$ and thereby obtain the posterior probability defined in Definition 3.1.

Given that $D_2$ is provided as a dataset and $\mathbb{I}(\cdot)$ denotes the indicator function, we can approximate the true probability using the empirical probability:
\begin{equation}
\small{
    f(y=r_k|x, D_2) \approx \hat{\mathbb{E}}(y=r_k|x, D_2) = \frac{\sum \mathbb{I}(y=r_k)}{\sum \mathbb{I}(y)} \Bigg|_{x, D_2}
}
\end{equation}

However, the token sequence space is high-dimensional and explicitly sparse; actual data points matching $x$ exactly are scarce. To mitigate data sparsity, we relax the exact match condition $x$ to a neighborhood of contexts $\tilde{x}$ similar to $x$:

\begin{equation}
    \label{similar_x}
\small{
    \frac{\sum \mathbb{I}(y=r_k)}{\sum \mathbb{I}(y)} \Bigg|_{x, D_2} \approx \frac{\sum_{\tilde{x}} \mathbb{I}(y=r_k \mid \tilde{x}) \cdot a(x, \tilde{x})}{\sum_{\tilde{x}} \sum_{y'} \mathbb{I}(y' \mid \tilde{x}) \cdot a(x, \tilde{x})}
}
\end{equation}
where $a(x, \tilde{x})$ is a weighting term based on similarity between $x$ and $\tilde{x}$.
This method can be regarded as an extension of the classical Kernel Density Estimation approach.

Based on the derivation above, assuming access to only the open-source weights of a pre-trained model and a new dataset, we can effectively approximate an ideal model retrained on the new data solely through the inference-time posterior probability correction. This approach can offer practical value for addressing a wide range of real-world challenges in domain and personalized adaptation.

\section{Method}
\label{sec:method}
As derived in Section \ref{sec:theory}, we can effectively adapt a pre-trained model to a specific domain or personalized scenario by correcting the posterior probability using the supplementary dataset $D_2$. Although the theoretical framework has provided a rigorous foundation, direct application still faces practical challenges regarding efficiency and robustness.
This section presents SteerWrite, a practical framework tailored to bridge this gap. We first outline the overall workflow, as illustrated in Figure \ref{fig:method_overview}, followed by the specific implementation details of retrieval and estimation strategies.

\subsection{Framework Overview}
\label{sec:overview}
The workflow of SteerWrite centers on an efficient, training-free autoregressive pipeline supported by a lightweight personalized dataset.
Before inference, the supplementary personalized dataset is encoded offline using the prefill mechanism of the base LLM, constructing a vector database that naturally aligns with the model's representation space without any gradient updates.

During the online inference, for each autoregressive token generation step, the input sequence is processed by the LLM core to yield the contextual hidden state of the current context.
This hidden state serves a dual purpose: it is simultaneously projected to the vocabulary space to form the base model's original posterior (depicted as the yellow distribution), and employed as a query vector to retrieve the most semantically relevant contexts from the external vector database.

Crucially, the raw retrieval is insufficient for stable steering; therefore, the retrieved candidates first undergo refinement via rank-based calibration and historical momentum integration.
These refined signals are then aggregated to estimate a probability density using a Kernel Density Estimation (KDE) approach to produce the final target-specific steering distribution (depicted as the green distribution).
Finally, this external probability is merged with the base model's original posterior to yield the mixture posterior, from which the next token is sampled.

\subsection{Efficient Similarity Computation}
\label{sec:similarity}
Language contains complex patterns that heuristic rules fail to capture, while employing an external deep neural network incurs significant computational overhead. To balance performance and efficiency, we leverage the pre-trained model itself to extract representations.

Specifically, for a sequence of length $L$, the model generates $L$ hidden states in a single forward pass (i.e., prefill phase). Each hidden state $h_t$ at position $t$ effectively encodes the full semantic trajectory of the prefix context $x_{1} \dots x_t$. We adopt this hidden state directly as the representation for the context ending at $t$. This prefill process is also executed during the normal LLM generation process, so no additional computation is required.

For supplementary dataset $D_2$, we treat every token position in every document as a potential context candidate $\tilde{x}$.
Consider a document in $D_2$ consisting of tokens $w_1 w_2 \dots w_L$. This document provides $L$ context-target pairs: for each step $t \in [1, L]$, sequence $\tilde{x} = w_1 \dots w_t$ is context and $y = w_{t+1}$ is target (where $w_{L+1}$ is the special \texttt{<EOS>} token).
Since the dataset is static, we can pre-compute the hidden states for all documents in one pass per document, regardless of sequence length. This results in a datastore whose number of representations equals the total number of tokens in $D_2$.

During inference, given the current input context $x$, we compute the similarity $a_{\text{raw}}(x, \tilde{x})$ between the current hidden state and each candidate $\tilde{x}$ in the supplementary dataset $D_2$. In practice, this similarity can be instantiated with either an $L_2$-based metric or cosine similarity. The resulting computation remains efficient for the small supplementary datasets considered in this work.

\subsection{Robust Estimation Strategy}
Directly using the raw similarities from a sparse dataset would introduce significant high-variance noise. To mitigate this, we employ the following three regularization techniques.

\paragraph{Rank-based Calibration.}
\label{sec:calibration}
Instead of setting a hard threshold for retrieval, we employ a \textit{Rank-based Calibration} mechanism that implicitly performs filtering. 
We define a fixed, highly sparse template distribution $\mathcal{B} = \{b_1, b_2, \dots, b_N\}$, where $N$ is the total number of tokens in $D_2$. 

The values in $\mathcal{B}$ follow a steep decay, and critically, only the top fraction (e.g., top 0.1\%) of values are non-zero.
We sort all candidates $\tilde{x}$ in $D_2$ based on their raw similarity $a_{\text{raw}}(x, \tilde{x})$ in descending order. The $j$-th ranked candidate is then assigned the calibrated weight $b_j$ from the template.
This strategy enforces a stable retrieval distribution for the top candidates and automatically discards the long tail of irrelevant contexts.

\paragraph{Temporal Momentum.}
\label{sec:momentum}

A critical failure mode when guiding generation via external datasets is the tendency to collapse into degenerate repetition loops. This stems from a self-reinforcing positive feedback mechanism: retrieving and generating a high-frequency token can shift the context state $h(x)$ closer to the representation of that token, effectively trapping the model in a local basin.

To enable the model to evolve continuously and fluently in a manner akin to natural language, we propose \textit{Temporal Momentum}, which optimizes the current weights $a^{(t)}$ with the smoothed weights from the previous time step $t-1$:
\begin{align}
        a_{\text{smooth}}^{(t)}(x_{<t}, \tilde{x}_{<j})&=(1-\lambda_1)\cdot a_{\text{cal}}^{(t)}(x_{<t}, \tilde{x}_{<j})\\
        & + \lambda_1\cdot a_{\text{smooth}}^{(t-1)}(x_{<t-1}, \tilde{x}_{<j-1})
    \label{eq:tm}
\end{align}
where $\tilde{x}_{<j-1}$ denotes the expected predecessor of $\tilde{x}_{<j}$ in the dataset trajectory. This momentum term acts as an inertial guide, encouraging the generation trajectory to exhibit the same sequential continuity found in natural language.

\paragraph{Frequency Scaling.}
\label{sec:scaling}

Relying on a single mechanism is often insufficient to fully prevent degeneration; thus, we introduce \textit{Frequency Scaling} to penalize the probability of high-frequency tokens:
\begin{equation}
    p_{\text{scale}}(y=r_k \mid x) \propto p_{\text{smooth}}(y=r_k \mid x) \cdot (N(r_k))^{-\lambda_2}
    \label{eq:scale}
\end{equation}
where $N(r_k)$ denotes the global count of token $r_k$ in dataset $D_2$, and $\lambda_2 > 0$ determines the penalty strength. 
By explicitly down-weighting globally frequent tokens, we encourage the model to attend to richer, more specific patterns within the text, thereby averting repetition loops.

\begin{table*}[htb!]
  \centering
  \vspace{-0.5em}
  \scalebox{0.8}{
    \begin{tabular}
    {>{\centering}m{1.5cm}
>{\centering}m{1.1cm}
>{\centering}m{1.1cm}
>{\centering}m{1.2cm}
>{\centering}m{1.1cm}
>{\centering}m{1.1cm}|
>{\centering}m{1.1cm}
>{\centering}m{1.1cm}
>{\centering}m{1.2cm}
>{\centering}m{1.1cm}
>{\centering\arraybackslash}m{1.1cm}
}
    \hline
    \hline
    \multirow{2}{*}{methods} & \multicolumn{5}{c|}{CodeNet}          & \multicolumn{5}{c}{HMR} \\
          & Lev. & key. &Jaccard&AWV& Qwen3 & Lev. & key. &Jaccard&AWV& Qwen3 \\
    \hline
    0.6B-Base & 42.76  & 15.97  & 36.02  & 73.16  & 71.94  & 30.94  & 16.08  & 30.19  & 72.66  & 73.83  \\
    prompt & 42.31  & 15.27  & 36.04  & 73.03  & 71.68  & 30.74  & 15.69  & 29.95  & 72.61  & 73.78  \\
    RAG   & 43.33  & 16.13  & 36.16  & 72.80  & 71.85  & 36.22  & 22.50  & 34.84  & 74.20  & 76.01  \\
    DenseRAG & 47.22  & 21.99  & 40.59  & 73.38  & 73.76  & 35.96  & 22.28  & 34.71  & 74.43  & 75.82  \\
    RankRAG & 49.12  & 24.73  & 42.03  & 74.22  & 74.45  & 34.82  & 20.80  & 33.45  & 74.02  & 75.60  \\
    kNN-LM & 54.66  & 33.64  & 49.15  & 77.94  & 78.02  & 37.46  & 25.73  & 36.53  & 76.27  & 77.23  \\
    CAD   & 44.79  & 17.98  & 36.00  & 71.58  & 71.50  & 35.17  & 21.14  & 33.63  & 73.80  & 75.49  \\
    SteerWrite & \textbf{77.64 } & \textbf{67.59 } & \textbf{75.07 } & \textbf{90.42 } & \textbf{86.93 } & \textbf{52.19 } & \textbf{43.51 } & \textbf{50.11 } & \textbf{80.85 } & \textbf{80.68 } \\
    \hline
    1.7B-Base & 44.98  & 19.30  & 38.17  & 74.03  & 74.17  & 34.55  & 22.21  & 34.20  & 75.83  & 76.24  \\
    prompt & 44.58  & 19.00  & 38.29  & 74.04  & 74.04  & 34.33  & 21.87  & 33.88  & 75.69  & 76.20  \\
    RAG   & 46.40  & 21.36  & 39.95  & 75.11  & 74.65  & 38.91  & 27.59  & 37.91  & 77.18  & 78.05  \\
    DenseRAG & 48.91  & 24.31  & 41.75  & 73.99  & 74.95  & 38.47  & 27.04  & 37.59  & 76.96  & 77.77  \\
    RankRAG & 50.66  & 27.21  & 43.79  & 75.26  & 75.83  & 37.73  & 26.02  & 36.77  & 76.94  & 77.62  \\
    kNN-LM & 55.05  & 34.02  & 49.33  & 78.57  & 78.21  & 39.50  & 28.67  & 38.41  & 76.66  & 77.99  \\
    CAD   & 46.74  & 20.56  & 38.16  & 72.50  & 72.91  & 37.77  & 25.95  & 36.71  & 76.05  & 77.36  \\
    SteerWrite & \textbf{80.00 } & \textbf{71.15 } & \textbf{77.82 } & \textbf{91.65 } & \textbf{88.51 } & \textbf{53.00 } & \textbf{45.16 } & \textbf{50.98 } & \textbf{80.19 } & \textbf{81.66 } \\
    \hline
    4B-Base & 47.87  & 24.04  & 42.09  & 75.41  & 75.99  & 36.95  & 26.64  & 36.81  & 77.66  & 77.72  \\
    prompt & 47.83  & 24.23  & 41.75  & 75.71  & 76.19  & 37.00  & 26.74  & 36.78  & 77.77  & 77.81  \\
    RAG   & 49.42  & 25.59  & 42.74  & 76.15  & 76.07  & 41.57  & 31.99  & 40.75  & 79.27  & 79.64  \\
    DenseRAG & 53.00  & 30.98  & 46.16  & 76.26  & 77.31  & 40.77  & 30.93  & 40.05  & 78.58  & 78.86  \\
    RankRAG & 53.39  & 31.46  & 46.33  & 76.15  & 77.81  & 40.26  & 30.84  & 39.78  & 78.90  & 79.16  \\
    kNN-LM & 57.39  & 37.22  & 51.92  & 79.41  & 79.78  & 43.93  & 35.07  & 42.60  & 79.43  & 79.72  \\
    CAD   & 50.72  & 26.82  & 43.16  & 74.58  & 75.46  & 39.83  & 29.87  & 39.15  & 78.06  & 78.85  \\
    SteerWrite & \textbf{82.36 } & \textbf{74.74 } & \textbf{80.55 } & \textbf{93.03 } & \textbf{90.08 } & \textbf{56.38 } & \textbf{50.11 } & \textbf{54.02 } & \textbf{82.98 } & \textbf{83.24 } \\
    \hline
    \hline
    \multirow{2}{*}{methods} & \multicolumn{5}{c|}{UER}              & \multicolumn{5}{c}{Law} \\
          & Lev. & key. & Jaccard & AWV   & Qwen3 & Lev. & key. & Jaccard & AWV   & Qwen3 \\
    \hline
    0.6B-Base & 16.30  & 3.74  & 17.02  & 67.09  & 74.79  & 43.52  & 42.44  & 40.90  & 74.70  & 81.83  \\
    prompt & 16.28  & 3.74  & 17.06  & 66.99  & 74.82  & 42.81  & 41.96  & 40.35  & 74.15  & 81.76  \\
    RAG   & 28.10  & 18.12  & 26.73  & 72.29  & 77.42  & 56.13  & 53.31  & 52.21  & 79.65  & 84.73  \\
    DenseRAG &  27.89 & 19.22 & 27.05 & 73.40 & 78.46  & 62.15  & 60.34  & 58.16  & 83.38  & 86.51  \\
    RankRAG & 23.96  & 14.50  & 23.92  & 71.57  & 77.32  & 59.75  & 58.02  & 55.71  & 82.20  & 85.75  \\
    kNN-LM & 22.40  & 12.51  & 22.84  & 70.16  & 77.00  & 52.68  & 52.30  & 49.49  & 78.48  & 84.53  \\
    CAD   & 26.44  & 15.77  & 24.94  & 71.46  & 76.47  & 46.45  & 40.78  & 43.30  & 73.12  & 79.15  \\
    SteerWrite & \textbf{40.53 } & \textbf{35.26 } & \textbf{38.82 } & \textbf{76.83 } & \textbf{81.27 } & \textbf{63.10 } & \textbf{63.37 } & \textbf{59.83 } & \textbf{84.71 } & \textbf{87.17 } \\
    \hline
    1.7B-Base & 19.69  & 9.98  & 19.65  & 70.01  & 76.95  & 53.03  & 51.62  & 49.54  & 78.33  & 84.43  \\
    prompt & 19.57  & 9.67  & 19.50  & 69.64  & 76.85  & 50.75  & 49.10  & 47.46  & 78.42  & 83.79  \\
    RAG   & 30.91  & 23.57  & 29.45  & \textbf{74.92 } & 79.49  & 59.92  & 57.66  & 56.40  & 82.77  & 85.90  \\
    DenseRAG &  30.63 & 23.75 & 29.40 & 75.50 & 79.90  & 64.72  & 63.48  & 61.07  & 85.07  & 87.35  \\
    RankRAG & 26.76  & 19.44  & 26.25  & 73.75  & 78.95  & 63.00  & 61.55  & 59.06  & 84.55  & 86.85  \\
    kNN-LM & 22.30  & 10.77  & 21.71  & 66.28  & 76.21  & 59.58  & 58.46  & 57.11  & 81.43  & 86.24  \\
    CAD   & 28.94  & 20.75  & 27.46  & 74.08  & 78.40  & 48.95  & 43.36  & 45.85  & 75.76  & 80.58  \\
    SteerWrite & \textbf{40.41 } & \textbf{31.90 } & \textbf{37.22 } & 72.73  & \textbf{79.84 } & \textbf{65.08 } & \textbf{64.78 } & \textbf{61.89 } & \textbf{85.86 } & \textbf{87.63 } \\
    \hline
    4B-Base & 21.55  & 14.37  & 22.48  & 72.50  & 78.45  & 60.49  & 60.21  & 57.01  & 83.96  & 86.60  \\
    prompt & 21.54  & 14.33  & 22.49  & 72.59  & 78.58  & 59.64  & 59.72  & 56.36  & 83.07  & 86.24  \\
    RAG   & 32.79  & 27.15  & 31.70  & 76.78  & 81.05  & 64.66  & 63.33  & 61.44  & 84.88  & 87.23  \\
    DenseRAG &  32.28 & 26.85 & 31.27 & 77.16 & 81.14  & \textbf{67.20 } & \textbf{66.37 } & \textbf{63.92 } & \textbf{86.27 } & \textbf{88.04 } \\
    RankRAG & 28.54  & 23.10  & 28.50  & 75.82  & 80.36  & 66.71  & 66.24  & 63.31  & 86.37  & 87.92  \\
    kNN-LM & 28.04  & 21.39  & 28.01  & 72.61  & 79.86  & 62.39  & 60.99  & 59.99  & 81.33  & 87.06  \\
    CAD   & 30.92  & 24.31  & 29.80  & 75.59  & 79.88  & 54.28  & 50.93  & 51.91  & 78.28  & 82.16  \\
    SteerWrite & \textbf{47.36 } & \textbf{42.70 } & \textbf{43.86 } & \textbf{79.73 } & \textbf{83.80 } & 65.12  & 63.71  & 62.23  & 83.38  & 87.68  \\
    \hline
    \hline
    \end{tabular}
    }
    \vspace{-0.3em}
      \caption{Performance comparison of SteerWrite against seven baseline methods across multiple datasets and metrics. The best results within each experimental group are highlighted in \textbf{bold}.}
    \vspace{-0.5em}
  \label{tab:main_results}
\end{table*}

\section{Experimental Settings}
\label{sec:experiments}

\subsection{Datasets}
We construct and employ four diverse datasets spanning different domains, including clinical medicine, law, and code. Specifically, the CodeNet dataset is derived from the Python portion of the CodeNet repository \cite{puri2021codenet}, which we have further filtered and cleaned to ensure quality. For the medical domain, we introduce HMR (Hypertension Medical Reports) and UER (Ultrasound Examination Reports), both curated from real-world hospital clinical workflows and organized for the first time in this work. The Law dataset consists of legal documents crawled from the China Judgments Online platform\footnote{\url{https://wenshu.court.gov.cn/}}. For each dataset, we perform a relatively even split into a test set and an external supplementary dataset for steering. Further details can be found in Appendix \ref{app:dataset}.

\subsection{Models, Baselines, and Evaluation Protocol}
All experiments are conducted using three specific base models from the Qwen3 series \cite{yang2025qwen3}: Qwen3-0.6B-Base, Qwen3-1.7B-Base, and Qwen3-4B-Base. We deliberately select these lightweight architectures to simulate realistic edge-side deployment scenarios. Furthermore, utilizing raw base models rather than instruction-tuned variants ensures that the underlying mechanism aligns strictly with next-token prediction, theoretically consistent with our posterior probability framework.
Greedy decoding is employed throughout to ensure deterministic output. 

We compare SteerWrite with seven training-free baselines, covering the vanilla generation, prompt-level conditioning, retrieval-augmented generation, and inference-time decoding adaptation. These baselines comprehensively include the original base model, zero-shot prompting, RAG, DenseRAG, RankRAG, $k$NN-LM \cite{khandelwal2019generalization}, and Context-Aware Decoding (CAD)~\cite{shi2024trusting}. The implementation details of these baselines are provided in Appendix~\ref{app:dataset}.

To simulate the human-AI co-writing scenario and measure the alignment with human intent, we employ an interleaved evaluation protocol where we place an evaluation point approximately every 10 tokens within the test documents. 
At each test point, the model generates a continuation, which is then compared against the subsequent ground truth text. 
We evaluate the subsequent 40 characters for the HMR, UER, and Law datasets, extending this window to 80 characters for the CodeNet dataset.

\subsection{Evaluation Metrics}

Our evaluation philosophy centers on quantifying the tangible utility of AI co-writing. To this end, we employ a comprehensive set of five metrics, categorized into efficiency gains and semantic alignment.

The primary goal of a co-writing assistant is to minimize the manual effort required to transform the model's suggestion into the user's intended text.
To quantify this efficiency, Levenshtein edit distance \cite{lcvenshtcin1966binary} and user keystrokes are employed.
While Levenshtein distance (namely Lev.) measures the standard modification cost, the keystroke metric (namely key.) further corrects for physical keyboard actions, providing a nuanced approximation of real-world typing effort.
We report the relative reduction in human editing effort attributable to the model’s assistance\footnote{A keystrokes score of 30 implies that the method reduces the user's keystrokes by approximately 30\%}.
A higher score in these metrics directly correlates to alleviating human labor in actual documentation workflows.

However, relying solely on rigid character-level matching may underestimate the model's help, given that natural language is inherently flexible and a generated continuation may differ in phrasing yet convey the identical semantic intent.
To mitigate this strictness and capture textual equivalence, we incorporate three complementary similarity metrics: Jaccard similarity \cite{jaccard1901distribution} to measure the lexical overlap of word sets,
Average Word Vector similarity (AWV) to compare the mean embeddings (given by the embedding layer of Qwen2.5-0.5B \cite{qwen2.5Qwen}) of word tokens, 
and the Qwen3-embedding-8B score \cite{zhang2025qwen3} (namely Qwen3) to leverage state-of-the-art semantic representations.
These metrics ensure that valid variations in expression are comprehensively acknowledged alongside editing efficiency.

\section{Experimental Results}
\subsection{Main Experiments}

To evaluate the performance of our proposed method, we conduct comprehensive testing comparing SteerWrite with seven training-free baselines across four diverse datasets and five evaluation metrics. The results are presented in Table \ref{tab:main_results}. 

The outcomes demonstrate that our method achieves the strongest overall performance across nearly all testing scenarios. 
Specifically, on the CodeNet, HMR, and UER datasets, SteerWrite demonstrates substantial improvements on the two critical metrics representing editing efficiency, often doubling the performance gains compared to the baselines.
This significant margin underscores the capability of the proposed method to reduce the manual text input required from humans in real-world tasks.

It is worth noting that the stronger retrieval mechanisms such as DenseRAG and RankRAG can improve over standard RAG in some cases, especially on retrieval-friendly datasets. 
However, their gains are not consistent across domains, model scales, and metrics, suggesting that prompt-level context injection remains insufficient for stable co-writing assistance.
In contrast, SteerWrite applies token-level posterior correction throughout generation, leading to more robust improvements.

Furthermore, we observe that our method, when paired with the Qwen3-0.6B-Base model, outperforms the stand-alone 4B-base model across all four datasets.
This suggests that the performance gains derived from our steering framework far exceed the benefits obtained simply by scaling up model parameters.
Moreover, although the Qwen3 technical report claims state-of-the-art results on standard benchmarks due to the extensive scaling of code data during pre-training \cite{yang2025qwen3}, applying our method still delivers a significant improvement on the Python code dataset.
This observation emphasizes the superiority of our approach and highlights the considerable potential of leveraging supplementary datasets for inference-time guidance.
Finally, the results across different metrics within the same dataset exhibit a high degree of correlation, verifying the consistency and rationality of our selected evaluation protocol.

\subsection{Generation Length Study}
\label{sec:length_experiment}
The proposed method is positioned for the co-writing scenario, where determining the appropriate length for a single model continuation is a pivotal question.
The shorter generation windows are typically easier for the model to predict accurately but necessitate frequent human-AI interactions.
On the contrary, the longer windows carry the risk of producing substantial text that deviates from the user's intent and requires extensive correction.
It is worth mentioning that an unrestricted or infinite generation budget corresponds to the traditional paradigm, where the model attempts to complete the entire remainder of the document in a single pass.

\begin{table}[htbp]
\centering
\scalebox{0.8}{
\begin{tabular}
{>{\centering}m{1.1cm}
>{\centering}m{1.1cm}
>{\centering}m{1.1cm}
>{\centering}m{1.2cm}
>{\centering}m{1.1cm}
>{\centering\arraybackslash}m{1.1cm}
}
\hline
\hline
length & Lev.  & key.  & Jaccard & AWV   & Qwen3 \\
\hline
      & \multicolumn{5}{c}{base} \\
\hline
20    & 21.78  & 7.91  & 18.51  & 62.92  & 75.84  \\
40    & 16.30  & 3.74  & 17.02  & 67.09  & 74.79  \\
80    & 12.33  & -0.73  & 16.29  & 69.94  & 73.46  \\
120   & 10.48  & -3.81  & 15.57  & 70.53  & 72.92  \\
\hline
\hline
      & \multicolumn{5}{c}{prompt} \\
\hline
20    & 21.76  & 7.99  & 18.60  & 62.72  & 75.81  \\
40    & 16.28  & 3.74  & 17.06  & 66.99  & 74.82  \\
80    & 12.33  & -0.42  & 16.49  & 70.15  & 73.59  \\
120   & 10.59  & -3.26  & 15.90  & 70.86  & 73.15  \\
\hline
\hline
      & \multicolumn{5}{c}{RAG} \\
\hline
20    & 36.51  & 25.90  & 31.93  & 69.84  & 79.62  \\
40    & 28.10  & 18.13  & 26.76  & 72.30  & 77.45  \\
80    & 20.35  & 10.02  & 22.53  & 74.48  & 74.66  \\
120   & 16.45  & 5.12  & 20.43  & 75.10  & 73.60  \\
\hline
\hline
      & \multicolumn{5}{c}{SteerWrite} \\
\hline
20    & 50.16  & 43.79  & 45.26  & 76.21  & 83.43  \\
40    & 40.53  & 35.26  & 38.82  & 76.83  & 81.27  \\
80    & 31.96  & 27.44  & 34.00  & 77.73  & 78.85  \\
120   & 27.90  & 23.40  & 32.42  & 78.34  & 78.45  \\
\hline
\hline
\end{tabular}
}
\caption{Analysis of the relationship between model performance and the generation length.}
\label{tab:length_study}
\end{table}

To explore this trade-off, we analyze the relationship between model performance and generation length.
This experiment is conducted using the 0.6B model on UER dataset.
To keep the main text compact, Table~\ref{tab:length_study} reports SteerWrite and three typical baselines, while the complete results with all baselines can be found in Appendix~\ref{app:length_full}.

The results indicate that model performance generally degrades as the generation length increases.
In particular, a negative keystroke score indicates that the model generates excessive unusable content, such that the cost of deleting it outweighs the benefit provided by the suggestion.
Meanwhile, the observed counter-intuitive increase in the AWV metric stems from its computation as an average of token embedding vectors.
Averaging over a larger set of irrelevant vectors tends to pull the resultant mean toward a generic center, artificially inflating similarity.
Therefore, we recommend using AWV for comparison only under fixed generation lengths.
Overall, these results validate that blindly pursuing long, single-turn completions is often suboptimal, supporting our premise that interactive co-writing with finer-grained and periodic human guidance is a superior strategy.

\subsection{Runtime Efficiency}
\label{sec:runtime}
Low latency is essential for stream-based co-writing, where the model should provide responsive suggestions rather than perform offline long-form generation. We therefore evaluate the runtime efficiency of SteerWrite against all baselines using two standard metrics: Time To First Token (TTFT) and Time Per Output Token (TPOT), both measured in milliseconds.

As shown in Table~\ref{tab:runtime}, SteerWrite introduces only a modest latency overhead. Across all tested model sizes, its TTFT remains below 47 ms and its TPOT remains below 30 ms per token, which is practical for interactive co-writing. The additional cost mainly comes from the similarity computation over the external datastore and the subsequent posterior correction, both of which are parallelizable and lightweight under the small-dataset setting considered in this work. These results show that SteerWrite substantially improves the editing-effort reduction while preserving the responsiveness required by real-world co-writing systems.

\begin{table}[htbp]
\centering
\scalebox{0.9}{
\begin{tabular}
{>{\centering}m{1.4cm}
>{\centering}m{0.8cm}
>{\centering}m{0.8cm}
>{\centering}m{0.8cm}
>{\centering}m{0.8cm}
>{\centering}m{0.8cm}
>{\centering\arraybackslash}m{0.8cm}
}
\hline
\hline
\multirow{2}[2]{*}{method} & \multicolumn{2}{c}{0.6B-Base} & \multicolumn{2}{c}{1.7B-Base} & \multicolumn{2}{c}{4B-Base} \\
\cmidrule{2-7}
& TTFT & TPOT & TTFT & TPOT & TTFT & TPOT \\
\hline
Base  & 26.5  & 20.6  & 25.0  & 20.3  & 31.3  & 24.2  \\
prompt & 25.7  & 20.7  & 24.2  & 21.5  & 34.3  & 25.6  \\
RAG   & 34.8  & 21.3  & 38.2  & 21.5  & 63.8  & 25.5  \\
DenseRAG & 23.3  & 19.1  & 25.4  & 20.2  & 33.5  & 25.7  \\
RankRAG & 34.5  & 19.9  & 36.2  & 20.9  & 54.4  & 24.3  \\
$k$NN-LM & 37.5  & 21.9  & 40.0  & 23.8  & 45.1  & 28.8  \\
CAD   & 50.0  & 40.6  & 52.2  & 40.5  & 84.7  & 53.5  \\
SteerWrite  & 37.0  & 23.6  & 38.5  & 23.5  & 46.7  & 29.3  \\
\hline
\hline
\end{tabular}
}
\caption{Runtime analysis in milliseconds}
\label{tab:runtime}
\end{table}

\begin{figure}[hbtp]
\vspace{-0.5em}
\centering
\includegraphics[width=0.98\linewidth]{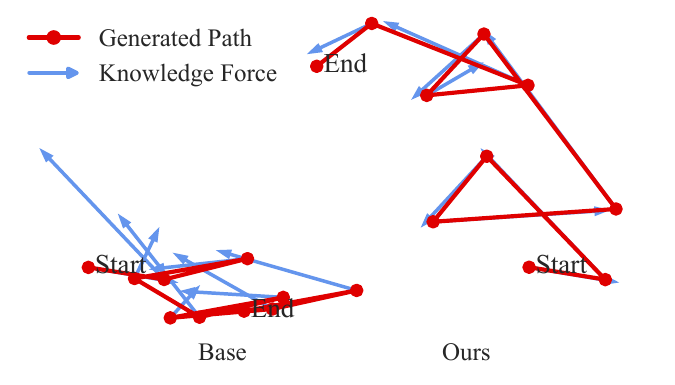}
\vspace{-0.5em}
\caption{Visualization of the model generation path and steering vector.}
\label{fig:viz}
\end{figure}

\subsection{Visualizing the Dynamics of Token-Level Steering}
\label{sec:viz_experiment}
To intuitively examine how SteerWrite affects generation, we visualize a case study using Qwen3-0.6B-Base. We project the hidden states at each decoding step into a 2D space with UMAP~\cite{mcinnes2018umap}, and connect them as a generation trajectory. At each step, we also project the retrieved token embeddings and aggregate them with the steering weights to form a "knowledge force" vector, indicating the direction suggested by the supplementary data. For clearer visualization, we set $\log \frac{p(D_2)}{p(D_1)}=1.0$.

As shown in Figure~\ref{fig:viz}, SteerWrite drives the generation trajectory to follow the knowledge-force direction, while the original base model, although starting from the same input context, quickly deviates from that direction. This contrast demonstrates that the external data guides generation continuously at the token level, rather than only shifting the initial state through prompt-level conditioning. The visualization therefore provides direct evidence for the necessity of token-level steering in the personalized co-writing scenario.

\section{Conclusion}
This work introduced SteerWrite, a training-free token-level framework designed to enable efficient, personalized co-writing for LLMs.
By leveraging a theoretical insight based on posterior probabilities and KDE, our proposed method effectively utilizes small-scale external datasets to steer the base model at the token level, avoiding the high costs associated with fine-tuning or coarse-grained prompt injection associated with RAG.
Extensive experiments across diverse domains demonstrated that SteerWrite achieved state-of-the-art performance, significantly outperforming larger base models, prompting methods, and standard RAG baselines in terms of both generation quality and human editing efficiency.
Furthermore, the visualization analysis confirmed the framework's capability to explicitly guide generation towards domain-specific distributions, validating the necessity of continuous steering over simple prompt initialization. Apart from methodology, this work also contributed the new benchmark datasets and evaluation framework for personalized co-writing. 

Looking ahead, we believe the personalized co-writing paradigm holds significant potential for broader applications. Although the proposed SteerWrite currently operates on textual data, a more valuable direction would be to investigate how such probability-based steering mechanisms can be extended to multimodal contexts, where the inputs, outputs, and reference datastores encompass images, audio, or video.
Addressing the alignment of heterogeneous modality representations for token-level guidance remains a challenging yet promising frontier for future research.

\appendix
\bibliography{aaai2027}

\clearpage

\section{Dataset and Experimental Setup Details}
\label{app:dataset}
Our work introduces four datasets from distinct domain scenarios, which have been briefly outlined in Section \ref{sec:experiments}. In this section, we provide additional descriptive details regarding their construction and characteristics.

For each dataset, we perform a random shuffle followed by a split into two subsets of equal document count. One subset serves as the held-out test set for evaluation, while the other functions as the external supplementary source used for retrieval in the RAG baseline and for steering in our SteerWrite framework.
We compute comprehensive descriptive statistics for these subsets, including the number of documents, total token count, and the average, minimum, and maximum token counts per document, as summarized in Table \ref{tab:dataset_stats}.

The supplementary portions are of a practical scale, containing 86.9k tokens for CodeNet, 42.6k tokens for HMR, 79.4k tokens for UER, and 90.2k tokens for Law.  
As indicated by the data, while there is significant variation in token length distributions across the four domains, the statistical characteristics between the two split subsets within each dataset remain highly consistent, ensuring a fair evaluation.
Notably, collecting and processing data at this scale is highly feasible for individual users, with the offline prefill phase typically completing within seconds on consumer-grade hardware.

\begin{table}[h!]
  \centering
  \scalebox{0.8}{
    \begin{tabular}{ccccccc}
    \hline
    \hline
          &       &\makecell{num of \\ doc.}& \makecell{num of \\ tokens}&\makecell{average \\ tokens} &\makecell{min \\ tokens} &\makecell{max \\ tokens} \\
    \hline
    \multirow{2}[2]{*}{CodeNet} & test  & 75    & 82102 & 1094.7  & 673   & 1805 \\
          & supp. & 75    & 86910 & 1158.8  & 804   & 2248 \\
    \hline
    \multirow{2}[2]{*}{HMR} & test  & 10    & 39773 & 3977.3  & 3122  & 5221 \\
          & supp. & 10    & 42603 & 4260.3  & 2790  & 5252 \\
    \hline
    \multirow{2}[2]{*}{UER} & test  & 100   & 81720 & 817.2  & 625   & 1295 \\
          & supp. & 100   & 79370 & 793.7  & 629   & 1424 \\
    \hline
    \multirow{2}[2]{*}{Law} & test  & 140   & 89431 & 638.8  & 341   & 1215 \\
          & supp. & 140   & 90157 & 644.0  & 373   & 1209 \\
    \hline
    \hline
    \end{tabular}
    }
      \caption{Descriptive statistics of the four domain datasets. The "test" and "supp." (supplementary) row denote the evaluation set and the supplementary reference set, respectively.}
  \label{tab:dataset_stats}
\end{table}

For the configuration of SteerWrite, we adopt a mostly fixed hyperparameter setting across the main experiments unless otherwise stated. Specifically, the logarithmic data ratio $\log\frac{p(D_2)}{p(D_1)}$ is set to 0.6, and the momentum decay factor $\lambda_1$ in Equation~\ref{eq:tm} is set to 0.5. For the raw similarity metric, we use an $L_2$-based similarity for Qwen3-0.6B-Base, and cosine similarity for Qwen3-1.7B-Base and Qwen3-4B-Base, according to the empirical behavior of different model representations. Correspondingly, the frequency scaling penalty $\lambda_2$ in Equation~\ref{eq:scale} is set to 0 for Qwen3-0.6B-Base, and 0.4 for Qwen3-1.7B-Base and Qwen3-4B-Base.

For the prompt-based baseline, we prepend a short task instruction to the user's writing prefix, with only the domain description adapted across datasets. For retrieval-based baselines, retrieval is performed independently at each test point rather than once per document, ensuring that the retrieved context is matched to the current writing prefix. Specifically, the standard RAG baseline retrieves the top-5 chunks from the supplementary subset using TF-IDF, while DenseRAG retrieves the top-5 chunks using Qwen3-Embedding-8B. RankRAG first retrieves 20 candidates with Qwen3-Embedding-8B and then selects the top-5 chunks using the BAAI bge-reranker-v2-m3 reranker. For CAD~\cite{shi2024trusting}, we use the RAG-retrieved contexts as the additional evidence.

\section{Computational Complexity Analysis}
\label{app:complexity}

We provide a complexity analysis of SteerWrite to clarify its storage and inference-time overhead. Let $N$ denote the number of tokens in the supplementary dataset and $d$ denote the hidden dimension of the base model. During the offline prefill stage, SteerWrite stores one hidden representation for each token position in the supplementary dataset. Therefore, the additional datastore size is $O(Nd)$. For a practical setting with $N=50{,}000$ and Qwen3-0.6B-Base ($d=1024$), the datastore contains $Nd = 50{,}000 \times 1024 = 51{,}200{,}000$
scalar entries, i.e., $0.0512$B parameter-equivalent entries. This corresponds to only about $7.7\%$ of the model parameter count. For Qwen3-4B-Base, this proportion further drops to $3.7\%$.

During inference, the base LLM decoding process remains unchanged. SteerWrite only introduces an additional token-level retrieval and posterior correction step. For each decoded token, the dominant extra operation is computing the similarity between the current hidden state and the $N$ stored representations. This introduces approximately $3Nd$ additional scalar operations per decoding step. Assuming a current sequence length of $1{,}000$ tokens, this overhead is roughly $13.6\%$ of the computation of Qwen3-0.6B-Base. For Qwen3-4B-Base, the relative overhead further decreases to $5.8\%$.

The empirical latency overhead in Section~\ref{sec:runtime} is slightly higher than the above theoretical computation ratio, especially for larger models. This gap is expected because our current implementation has not been heavily optimized and still includes practical overhead from datastore access, similarity computation, calibration, and framework-level operations. Meanwhile, the core additional computation is highly parallelizable and does not require any gradient update or input-context expansion. These results suggest that SteerWrite is already practical for interactive co-writing, while its runtime efficiency can be further improved through optimized datastore kernels and more efficient retrieval implementations.

\section{Generalization to Different Model Families}
\label{app:llama}

\begin{table}[h]
\centering
\scalebox{0.8}{
\begin{tabular}
{>{\centering}m{1.3cm}
>{\centering}m{1.3cm}
>{\centering}m{1.4cm}
>{\centering}m{1.3cm}
>{\centering}m{1.2cm}
>{\centering\arraybackslash}m{1.3cm}
}
\hline
\hline
method & Lev.  & key. & Jaccard & AWV   & Qwen3 \\
\hline
\multicolumn{6}{c}{Llama-3.2-1B} \\
Base  & 28.66 & -4.58 & 22.13 & 57.71 & 71.68 \\
prompt & 29.03 & -4.89 & 21.40  & 57.20  & 72.31 \\
RAG   & 31.21 & -0.84 & 23.64 & 58.68 & 72.41 \\
SteerWrite  & \textbf{56.74} & \textbf{52.72} & \textbf{61.97} & \textbf{81.79} & \textbf{76.87} \\
\hline
\multicolumn{6}{c}{Llama-3.2-3B} \\
Base  & 31.07 & -2.60  & 23.11 & 56.39 & 66.98 \\
prompt & 32.06 & -2.94 & 23.14 & 55.04 & 66.90 \\
RAG   & 32.26 & 0.42  & 24.86 & 57.15 & 68.16 \\
SteerWrite  & \textbf{57.29} & \textbf{53.18} & \textbf{62.98} & \textbf{82.29} & \textbf{77.23} \\
\hline
\hline
\end{tabular}
}
\caption{Llama series models}
\label{tab:llama}
\end{table}

To verify that SteerWrite is not specific to the Qwen3 model family, we conduct an additional experiment using the Llama-3.2 series~\cite{meta2024llama32}. Since Llama models have limited support for Chinese text generation, we perform this evaluation on the CodeNet dataset. We compare SteerWrite with representative baselines, including the original base model, prompt, and standard RAG. All evaluation metrics follow the same protocol as the main experiments.

As shown in Table~\ref{tab:llama}, SteerWrite consistently outperforms all representative baselines on both Llama-3.2-1B and Llama-3.2-3B. The improvements are especially large on the editing-effort metrics, where the base, prompt, and RAG baselines even yield negative or near-zero keystroke reductions, while SteerWrite achieves over 50 points on both model sizes. These results indicate that the proposed token-level steering mechanism is not tied to a specific model family and can generalize to different LLM architectures.

\section{Complete Results of Generation Length Study}
\label{app:length_full}

We provide the complete generation length study in Table~\ref{tab:length_study_full}, including all baselines evaluated in the main experiments.
The experiment is conducted on the UER dataset with Qwen3-0.6B-Base, following the same evaluation protocol as Section~\ref{sec:length_experiment}.

\begin{table}[htbp]
\centering
\scalebox{0.8}{
\begin{tabular}
{>{\centering}m{1.1cm}
>{\centering}m{1.1cm}
>{\centering}m{1.1cm}
>{\centering}m{1.2cm}
>{\centering}m{1.1cm}
>{\centering\arraybackslash}m{1.1cm}
}
\hline
\hline
length & Lev.  & key.  & Jaccard & AWV   & Qwen3 \\
\hline
\multicolumn{6}{c}{base} \\
\hline
20    & 21.78  & 7.91  & 18.51  & 62.92  & 75.84  \\
40    & 16.30  & 3.74  & 17.02  & 67.09  & 74.79  \\
80    & 12.33  & -0.73  & 16.29  & 69.94  & 73.46  \\
120   & 10.48  & -3.81  & 15.57  & 70.53  & 72.92  \\
\hline
\hline
\multicolumn{6}{c}{prompt} \\
\hline
20    & 21.76  & 7.99  & 18.60  & 62.72  & 75.81  \\
40    & 16.28  & 3.74  & 17.06  & 66.99  & 74.82  \\
80    & 12.33  & -0.42  & 16.49  & 70.15  & 73.59  \\
120   & 10.59  & -3.26  & 15.90  & 70.86  & 73.15  \\
\hline
\hline
\multicolumn{6}{c}{RAG} \\
\hline
20    & 36.51  & 25.90  & 31.93  & 69.84  & 79.62  \\
40    & 28.10  & 18.13  & 26.76  & 72.30  & 77.45  \\
80    & 20.35  & 10.02  & 22.53  & 74.48  & 74.66  \\
120   & 16.45  & 5.12  & 20.43  & 75.10  & 73.60  \\
\hline
\hline
\multicolumn{6}{c}{DenseRAG} \\
\hline
20    & 36.00  & 25.99  & 31.34  & 70.37  & 80.17  \\
40    &  27.89 & 19.22 & 27.05 & 73.40 & 78.46 \\
80    & 20.89  & 12.24  & 24.13  & 75.98  & 76.28  \\
120   & 17.30  & 7.63  & 22.40  & 76.69  & 75.48  \\
\hline
\hline
\multicolumn{6}{c}{RankRAG} \\
\midrule
20    & 31.32  & 20.48  & 27.38  & 67.93  & 78.84  \\
40    & 23.96  & 14.50  & 23.92  & 71.57  & 77.32  \\
80    & 17.86  & 8.29  & 21.67  & 74.44  & 75.37  \\
120   & 14.97  & 4.47  & 20.47  & 75.27  & 74.76  \\
\hline
\hline
\multicolumn{6}{c}{kNN-LM} \\
\hline
20    & 30.95  & 20.47  & 27.14  & 68.24  & 78.62  \\
40    & 22.40  & 12.51  & 22.84  & 70.16  & 77.00  \\
80    & 16.20  & 5.27  & 20.32  & 71.60  & 75.02  \\
120   & 13.48  & 1.04  & 18.88  & 71.70  & 74.42  \\
\hline
\hline
\multicolumn{6}{c}{CAD} \\
\hline
20    & 35.05  & 23.84  & 30.61  & 68.86  & 78.85  \\
40    & 26.44  & 15.77  & 24.94  & 71.46  & 76.47  \\
80    & 18.70  & 7.44  & 20.36  & 74.03  & 73.20  \\
120   & 14.95  & 2.50  & 18.28  & 74.89  & 72.02  \\
\hline
\hline
\multicolumn{6}{c}{SteerWrite} \\
\hline
20    & 50.16  & 43.79  & 45.26  & 76.21  & 83.43  \\
40    & 40.53  & 35.26  & 38.82  & 76.83  & 81.27  \\
80    & 31.96  & 27.44  & 34.00  & 77.73  & 78.85  \\
120   & 27.90  & 23.40  & 32.42  & 78.34  & 78.45  \\
\hline
\hline
\end{tabular}
}
\caption{Analysis of the relationship between model performance and the generation length.}
\label{tab:length_study_full}
\end{table}

The results are consistent with the findings in the main text: longer continuations generally lead to lower editing-effort reduction, especially on the keystroke metric.
This confirms that overly long single-turn completions are often less suitable for interactive co-writing, where short and periodic suggestions better preserve alignment with the user's intended text.

\section{Data Ratio Analysis}
\label{sec:ratio_analysis}
A critical hyperparameter of the proposed method is the data ratio term presented in Equation (\ref{pxmain}). 
Conceptually, this ratio reflects the prior belief regarding the source of the generated content before any token generation begins.
In other words, it quantifies the extent to which the specific domain of the current task was represented in the model's original pre-training corpus.

When the extensive knowledge acquired during pre-training significantly exceeds the small-scale supplementary dataset, implying $p(D_1) \gg p(D_2)$, the ratio $\frac{p(D_2)}{p(D_1)} \approx 0$, causing our method to degenerate to relying entirely on the model's internal distribution.
Conversely, if the specialized domain is rarely encountered during the pre-training phase, this ratio naturally increases to favor the external distribution.
However, even in cases where the domain appears entirely novel, we advise against setting an excessively large ratio, as over-reliance on the sparse external dataset can lead to severe instability in the generation process.
This constraint implies that regardless of the domain specificity, any text inevitably shares basic linguistic similarities, such as basic grammatical rules, with general corpora, preserving partial applicability and value of the pre-trained knowledge.

To validate the above analysis, we conduct experiments examining the relationship between model performance and the data ratio.
We employ the three models with different parameter sizes and the UER dataset, setting the log data ratio $\log\frac{p(D_2)}{p(D_1)}$ to values of 0.0, 0.2, 0.4, 0.6, 0.8, and 1.0.
The log scale is employed because the internal computations of LLMs operate primarily within the log-probability space.
The experimental results are presented in Table \ref{tab:ratio_study}.

\begin{table}[htb!]
  \centering
    \scalebox{0.8}{
    \begin{tabular}
    {>{\centering}m{1.2cm}
>{\centering}m{1.1cm}
>{\centering}m{1.4cm}
>{\centering}m{1.2cm}
>{\centering}m{1.1cm}
>{\centering\arraybackslash}m{1.8cm}
}
    
\hline
\hline
log ratio  & Lev. & key. & Jaccard & AWV   & Qwen3 \\
\hline
\multicolumn{6}{c}{Qwen3-0.6B-Base} \\
\hline
0.0   & 36.73  & 30.57  & 35.04  & 75.73  & 80.32  \\
0.2   & 39.30  & 33.65  & 37.55  & 76.61  & 80.93  \\
0.4   & 40.07  & 34.72  & 38.39  & 76.82  & 81.16  \\
0.6   & 40.53  & 35.26  & 38.82  & 76.83  & 81.27  \\
0.8   & \textbf{40.65 } & \textbf{35.45 } & \textbf{38.98 } & \textbf{76.84 } & \textbf{81.34 } \\
1.0   & 40.49  & 35.29  & 38.88  & 76.64  & 81.31  \\
\hline
\hline
\multicolumn{6}{c}{Qwen3-1.7B-Base} \\
\hline
0.0   & 37.58  & 30.30  & 34.75  & \textbf{74.76 } & \textbf{80.49 } \\
0.2   & 39.32  & 31.68  & 36.50  & 74.64  & 80.41  \\
0.4   & \textbf{40.49 } & \textbf{32.59 } & \textbf{37.39 } & 73.81  & 80.33  \\
0.6   & 40.41  & 31.90  & 37.22  & 72.73  & 79.84  \\
0.8   & 40.35  & 31.16  & 36.99  & 71.60  & 79.33  \\
1.0   & 39.91  & 30.09  & 36.59  & 70.32  & 78.77  \\
\hline
\hline
\multicolumn{6}{c}{Qwen3-4B-Base} \\
\hline
0.0   & 43.63  & 39.21  & 40.89  & 79.34  & 83.30  \\
0.2   & 45.25  & 40.76  & 42.33  & 79.48  & 83.63  \\
0.4   & 46.78  & 42.34  & 43.44  & \textbf{79.84 } & 83.79  \\
0.6   & \textbf{47.36 } & \textbf{42.70 } & \textbf{43.86 } & 79.73  & \textbf{83.80 } \\
0.8   & 47.18  & 42.11  & 43.70  & 79.14  & 83.42  \\
1.0   & 46.60  & 41.06  & 43.20  & 78.30  & 83.09  \\
\hline
\hline
\end{tabular}
  }
    \caption{Analysis of method performance with different data ratios.}
\label{tab:ratio_study}
\end{table}

As anticipated, the results reveal that as the log ratio increases,  the method's performance initially improves and then subsequently declines.
This trajectory aligns perfectly with our analysis,
confirming that an optimal balance between internal pre-trained priors and external steering signals is essential for maximizing generation quality.

Furthermore, increasing the model parameter size does not lead to a significant reduction in the optimal configuration for the data ratio. 
This suggests that the specific knowledge contained within our supplementary dataset is indeed extremely rare across the pre-training distributions of the models.
Consequently, this insight provides a crucial indirect validation explaining why the 0.6B model equipped with our framework can outperform the unassisted 4B model in the main experiments, as the targeted external steering effectively compensates for specific knowledge gaps that simple parameter scaling fails to address.

\section{Ablation Study}
\label{app:ablation}

We conduct an ablation study to examine the contributions of two key stabilization components in SteerWrite: rank-based calibration and temporal momentum. Rank-based calibration is designed to transform raw similarity scores into a stable and sparse external distribution, while temporal momentum encourages continuity across adjacent decoding steps. To evaluate whether these components are consistently effective across model scales, we conduct experiments on the UER dataset using all three Qwen3 base models. The results are shown in Table~\ref{tab:ablation_study}.

\begin{table}[htbp]
  \centering
\scalebox{0.8}{
\begin{tabular}
{>{\centering}m{2.3cm}
>{\centering}m{1.1cm}
>{\centering}m{1.1cm}
>{\centering}m{1.2cm}
>{\centering}m{1.1cm}
>{\centering\arraybackslash}m{1.1cm}
}
\hline
\hline
method & Lev.  & key.  & Jaccard & AWV   & Qwen3 \\
\hline
\multicolumn{6}{c}{Qwen3-0.6B-Base} \\
\hline
SteerWrite & \textbf{40.53 } & \textbf{35.26 } & \textbf{38.82 } & \textbf{76.83 } & \textbf{81.27 } \\
w/o calibration & 36.66  & 30.32  & 35.40  & 74.90  & 79.88  \\
w/o momentum & 33.33  & 26.59  & 33.06  & 72.91  & 79.50  \\
\hline
\multicolumn{6}{c}{Qwen3-1.7B-Base} \\
\hline
SteerWrite & \textbf{40.41 } & \textbf{31.90 } & \textbf{37.22 } & \textbf{72.73 } & \textbf{79.84 } \\
w/o calibration & 37.18  & 28.35  & 34.62  & 71.58  & 79.08  \\
w/o momentum & 30.67  & 14.83  & 28.54  & 59.33  & 73.65  \\
\hline
\multicolumn{6}{c}{Qwen3-4B-Base} \\
\hline
SteerWrite & \textbf{47.36 } & \textbf{42.70 } & \textbf{43.86 } & \textbf{79.73 } & \textbf{83.80 } \\
w/o calibration & 43.16  & 37.34  & 40.60  & 77.38  & 82.48  \\
w/o momentum & 34.02  & 20.26  & 31.51  & 66.38  & 77.51  \\
\hline
\hline
\end{tabular}
}
  \caption{Ablation study on key stabilization components of SteerWrite across different model sizes.}
  \label{tab:ablation_study}
\end{table}

The results show that removing either component consistently degrades performance across all three model sizes. Removing rank-based calibration leads to clear declines on all metrics, indicating that directly using raw similarity scores is insufficient for deriving a reliable steering distribution from the external datastore. The degradation caused by removing temporal momentum is even more pronounced, especially on the 1.7B and 4B models, where the keystroke score drops substantially. This suggests that temporal continuity is crucial for stable token-level steering during autoregressive generation.

\begin{figure*}[htb!]
  \centering
  \includegraphics[width=500pt]{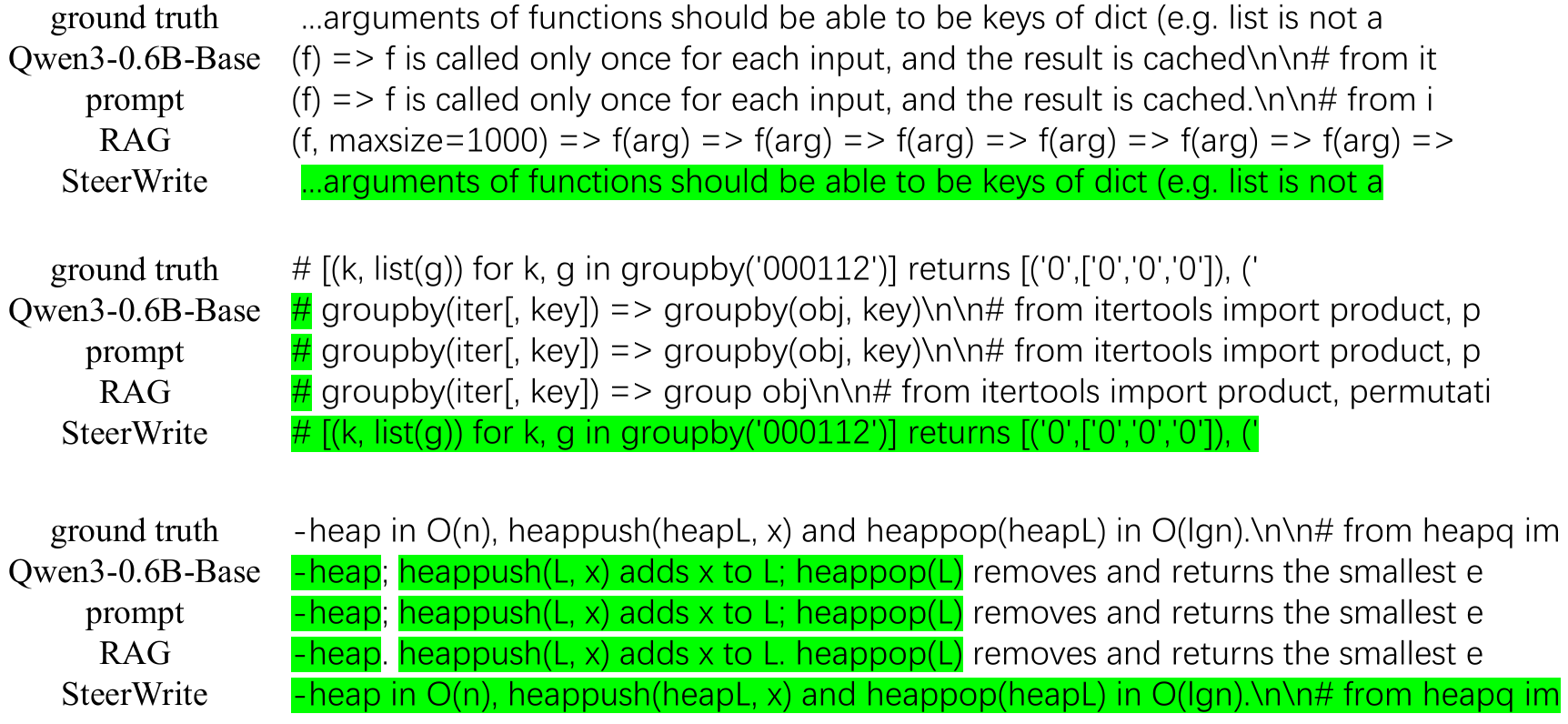}
  \caption{Intuitive Examples of co-writing completions on CodeNet dataset. Text segments identical to the ground truth are highlighted in green.}
  \label{fig:case_study1}
\end{figure*}

Overall, the ablation results confirm that both rank-based calibration and temporal momentum are necessary for robust posterior correction. Calibration improves the quality of the external distribution at each decoding step, while momentum stabilizes the generation trajectory across steps. Their combined effect enables SteerWrite to maintain consistent improvements across different model scales.

\section{Intuitive Examples of Co-Writing Completions}
\label{app:intuitive}
\begin{figure*}[htb!]
  \centering
  \includegraphics[width=500pt]{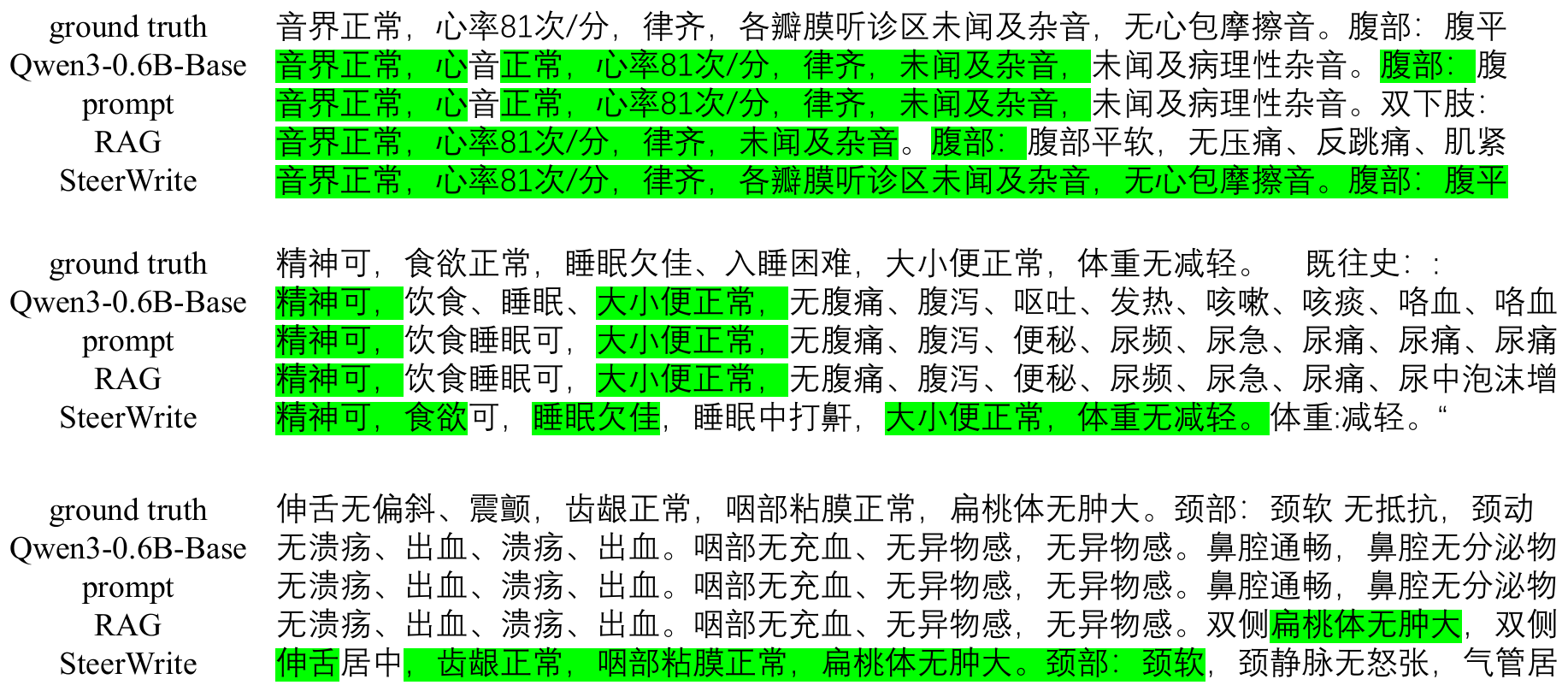}
  \caption{Intuitive Examples of co-writing completions on HMR dataset. Text segments identical to the ground truth are highlighted in green.}
  \label{fig:case_study2}
\end{figure*}

To intuitively demonstrate the effectiveness of our method, we select three representative test points from the CodeNet and HMR datasets.
For each instance, we present the reference ground truth alongside the co-writing continuations generated by SteerWrite and the three baseline methods.
To enhance visual clarity and facilitate immediate comparison, text segments that align verbatim with the ground truth are highlighted in green.
The comparative results are visualized in Figure \ref{fig:case_study1} and \ref{fig:case_study2}.

As we can observe from these examples, the continuations generated by SteerWrite exhibit more matching texts compared to the baseline methods.
These observations confirm that SteerWrite produces content much closer to the user's actual intent, empirically validating the substantial reduction in human editing effort provided by our method in real-world workflows.

\end{document}